\documentclass[conference]{IEEEtran}
\IEEEoverridecommandlockouts
\usepackage{cite}
\usepackage{amsmath,amssymb,amsfonts}
\usepackage{algorithmic}
\usepackage{graphicx}
\usepackage{flushend}
\usepackage{hyperref}
\usepackage{textcomp}
\usepackage{xcolor}
\usepackage{mwe}
\usepackage{subfig}
\usepackage[justification=centering]{caption}
 \usepackage{booktabs}

\def\BibTeX{{\rm B\kern-.05em{\sc i\kern-.025em b}\kern-.08em
    T\kern-.1667em\lower.7ex\hbox{E}\kern-.125emX}}
    
\newcommand{\ra}[1]{\renewcommand{\arraystretch}{#1}}

\begin{document}

\title{Squeezed Very Deep Convolutional Neural Networks for Text Classification
{}
\thanks{*Authors contributed equally and are both first writers.}
}

\author{
\IEEEauthorblockN{Andr\'ea B. Duque\footnotemark*}
\IEEEauthorblockA{\textit{Centro} \textit{de} \textit{Inform\'{a}tica} \\
\textit{Universidade} \textit{Federal} \textit{de} \textit{Pernambuco}\\
50.740-560, Recife, PE, Brazil\\
abd@cin.ufpe.com}
\\
\IEEEauthorblockN{David Mac\^edo}
\IEEEauthorblockA{\textit{Centro} \textit{de} \textit{Inform\'{a}tica} \\
\textit{Universidade} \textit{Federal} \textit{de} \textit{Pernambuco}\\
50.740-560, Recife, PE, Brazil\\
dlm@cin.ufpe.br}
\and
\IEEEauthorblockN{Lu\~a L\'azaro J. Santos\footnotemark*}
\IEEEauthorblockA{\textit{Centro} \textit{de} \textit{Inform\'{a}tica} \\
\textit{Universidade} \textit{Federal} \textit{de} \textit{Pernambuco}\\
50.740-560, Recife, PE, Brazil\\
lljs@cin.ufpe.br}
\\
\IEEEauthorblockN{Cleber Zanchettin}
\IEEEauthorblockA{Centro de Inform\'{a}tica}
\textit{Universidade} \textit{Federal} \textit{de} \textit{Pernambuco}\\
50.740-560, Recife, PE, Brazil\\
cz@cin.ufpe.br
}

\maketitle

\begin{abstract}
Most of the research in convolutional neural networks has focused on increasing network depth to improve accuracy, resulting in a massive number of parameters which restricts the trained network to platforms with memory and processing constraints. We propose to modify the structure of the Very Deep Convolutional Neural Networks (VDCNN) model to fit mobile platforms constraints and keep performance. In this paper, we evaluate the impact of Temporal Depthwise Separable Convolutions and Global Average Pooling in the network parameters, storage size, and latency.  The squeezed model (SVDCNN) is between 10x and 20x smaller, depending on the network depth, maintaining a maximum size of 6MB. Regarding accuracy,  the network experiences a loss between 0.4\% and 1.3\% and obtains lower latencies compared to the baseline model.


\end{abstract}

\medskip

\vspace{-3mm}

\section{Introduction}
The general trend in deep learning approaches has been developing models with increasing layers. Deeper neural networks have achieved high-quality results in different tasks such as image classification, detection, and segmentation. Deep models can also learn hierarchical feature representations from images \cite{zeiler2014visualizing}. In the Natural Language Processing (NLP) field, the belief that compositional models can also be used to text-related tasks is more recent.

The increasing availability of text data motivates research for models able to improve accuracy in different language tasks. Following the image classification Convolutional Neural Network (CNN) tendency, the research in text classification has placed effort into developing deeper networks. The first CNN based approach for text was a shallow network with one layer \cite{kim2014convolutional}. Following this work, deeper architectures were proposed \cite{conneau2016very,zhang2015character}. Conneau et al. \cite{conneau2016very} were the first to propose Very Deep Convolutional Neural Networks (VDCNN) applied to text classification. VDCNN accuracy increases with depth. The approach with 29 layers is the state-of-the-art accuracy of CNNs for text classification.

However, regardless of making networks deeper to improve accuracy, little effort has been made to build text classification models to constrained resources. It is a very different scenario compared to image approaches, where size and speed constrained models have been proposed \cite{iandola2016squeezenet, howard2017mobilenets}. In real-world applications, size and speed are constraints to an efficient mobile and embedded deployment of deep models \cite{howard2017mobilenets}.

Several relevant real-world applications depend on text classification tasks such as sentiment analysis, recommendation and opinion mining. The appeal for these applications combined with the boost in mobile devices usage motivates the need for research in restrained text classification models. Concerning mobile development, there are numerous benefits to developing smaller models. Some of the most relevant are requiring fewer data transferring while updating the client model \cite{iandola2016squeezenet} and increasing usability by diminishing the inference time. Such advantages would boost the usage of deep neural models in text-based applications for embedded platforms.

In this paper, we investigate modifications on the network proposed by Conneau et al. \cite{conneau2016very} with the aim of reducing its number of parameters, storage size and latency with minimal performance degradation. To achieve these improvements we used Temporal Depthwise Separable Convolution and Global Average Pooling techniques.
Therefore, our main contribution is to propose the Squeezed Very Deep Convolutional Neural
Networks (SVDCNN), a text classification model which requires significantly fewer parameters compared to the state-of-the-art CNNs.

Section II provides an overview of the state-of-the-art in CNNs for text classification. Section III presents the VDCNN model. Section IV explains the proposed model SVDCNN and the subsequent impact in the total number of parameters of the network. Section V details how we perform experiments. Section VI analyses the results and lastly, Section VII, presents conclusions and direction for future works.

\section{Related Work}
CNNs were originally designed for Computer Vision with the aim of considering feature extraction and classification as one task \cite{lecun1998gradient}. Although CNNs are very successful in image classification tasks, its use in text classification is relatively new and has some peculiarities. Contrasting with traditional image bi-dimensional representations, texts are one-dimensionally represented. Due to this property, the convolutions are designed as temporal convolutions. Furthermore, it is necessary to generate a numerical representation from the text so the network can be trained using this representation. This representation, namely embeddings, is usually obtained through the application of a lookup table, generated from a given dictionary.

An early approach for text classification tasks consisted of a shallow neural network working on the word level and using only one convolutional layer \cite{kim2014convolutional}. The author reported results in smaller datasets.  Later, Zhang et al. \cite{zhang2015character} proposed the first CNN performing on a character level (Char-CNN), which allowed them to train up to 6 convolutional layers, followed by three fully connected classification layers. Char-CNN uses convolutional kernels of size 3 and 7, as well as simple max-pooling layers. 

Conneau et al. (2016) proposed the Very Deep CNN (VDCNN) \cite{conneau2016very} also on a character level, presenting improvements compared to Char-CNN. Conneau et al. (2016) have shown that text classification accuracy increases when the proposed model becomes deeper. VDCNN uses only small kernel convolutions and pooling operations. The proposed architecture relies on the VGG and ResNet philosophy \cite{he2016deep,simonyan2014very}: The number of feature maps and the temporal resolution is modeled so that their product is constant.  This approach makes it easier to control the memory footprint of the network. Both Zhang and Conneau et al. CNNs utilized standard convolutional blocks and fully connected layers to combine convolution information \cite{zhang2015character, conneau2016very}. This architecture choice increases the number of parameters and storage size of the models. However, size and speed was not the focus of those works. 

The idea of developing smaller and more efficient CNNs without losing representative accuracy is a less explored research direction in NLP, but it has already been a trend for computer vision applications \cite{howard2017mobilenets,iandola2016squeezenet,santos2018reducing }.  Most approaches consist in compressing pre-trained networks or training small networks directly \cite{howard2017mobilenets}. A recent tendency in deep models is replacing standard convolutional blocks with Depthwise Separable Convolutions (DSCs). The purpose is to reduce the number of parameters and consequently the model size. DSCs were initially introduced in \cite{sifre2014rigid} and since then have been successfully applied to image classification and  \cite{howard2017mobilenets,santos2018reducing,chollet2017xception} machine translation \cite{kaiser2017depthwise} to reduce the computation in convolutional blocks.  Another approach is the use of a Global Average Pooling (GAP) layer at the output of the network to replace fully connected layers. This approach has become a standard architectural decision for newer CNNs \cite{he2016deep,huang2017densely}.

\begin{figure}[htbp]
\centering
\includegraphics[scale=.1042]{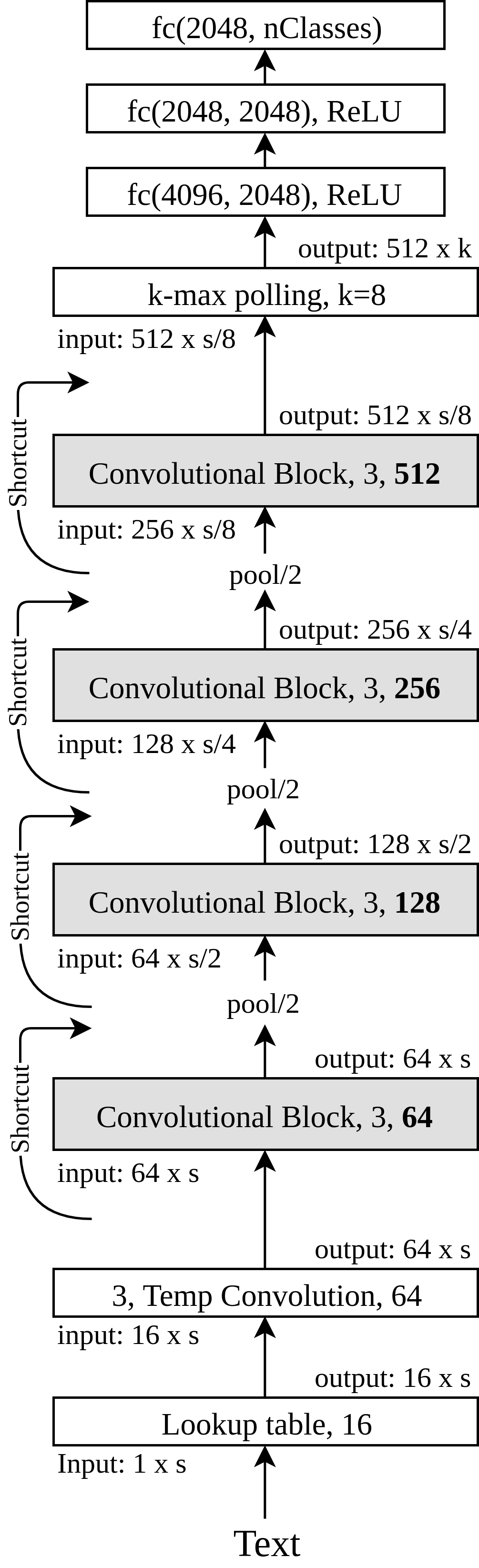}
\caption{Depth 9 VDCNN architecture.}
\label{vdcnn}
\end{figure}

\section{VDCNN Model for Text Classification}
The VDCNN is a modular architecture for text classification tasks developed to offer different depth levels (9, 17, 29 and 49). Fig.~\ref{vdcnn} presents the architecture for depth 9. The network begins with a lookup table, which generates the embeddings for the input text and stores them in a 2D tensor of size $(f0, s)$. The number of input characters $(s)$ is fixed to 1,024 while the embedding dimension $(f0)$ is 16. The embedding dimension can be seen as the number of RGB channels of an image. 

The following layer (3, Temp Convolution, 64) applies 64 temporal convolutions of kernel size 3, so the output tensor has size $64 \ast s$. Its primary function is to fit the lookup table output with the modular network segment input composed by convolutional blocks. Each aforenamed block is a sequence of two temporal convolutional layers,  each one accompanied by a temporal batch normalization layer \cite{ioffe2015batch} and a ReLU activation. Besides, the different network depths are obtained varying the number of convolutional blocks. As a convention,  the depth of a network is given as its total number of convolutions. For instance, the architecture of depth 17 has two convolutional blocks of each level of feature maps, which results in 4 convolutional layers for each level (see Table \ref{depth_vdcnn}).

\begin{table}[htpb]
\caption{Number of convolutional layers\\for each different VDCNN depth architecture}
\label{depth_vdcnn}
\ra{1.3}
\begin{center}
\begin{tabular}{@{}lrrrr@{}}\toprule
Depth          & 9 & 17 & 29 & 49  \\  \midrule
Convolutional Block 512 & 2 & 4 & 4 & 6 \\
Convolutional Block 256 & 2 & 4 & 4 & 10 \\
Convolutional Block 128 & 2 & 4 & 10 & 16 \\
Convolutional Block 64 & 2 & 4 & 10 & 16 \\
First Convolutional Layer & 1 & 1 & 1 & 1 \\
\bottomrule
\end{tabular}
\end{center}
\end{table}

\pagebreak

\noindent 
Considering the first convolutional layer of the network, we obtain the depth $2 \ast (2 + 2 + 2 +2)  + 1 = 17$. The different depth architectures provided by VDCNN model are summarized in Table \ref{depth_vdcnn}. The following rule is employed to minimize the network's memory footprint: Before each convolutional block doubling the number of feature maps, a pooling layer halves the temporal dimension. This strategy is inspired by the VGG and ResNets philosophy and results in three levels of feature maps: 128, 256 and 512 (see Fig.~\ref{vdcnn}). Additionally, the VDCNN network also contains shortcut connections  \cite{he2016deep} for each convolutional blocks implemented through the usage of $1\times1$ convolutions.

Lastly, for the classification task, the $k$ most valuable features $(k=8)$ are extracted using $k$-max pooling, generating a one-dimensional vector which supplies three fully connected layers with ReLU hidden units and softmax outputs. The number of hidden units is 2,048, and they do not use dropout but rather batch normalization after convolutional layers perform the network regularization.
\medskip

\section{SVDCNN Model for Text Classification}
\medskip
The primary objective is reducing the number of parameters so that the resulting network has a significative lower storage size. We first propose to modify the convolutional blocks of VDCNN model by the usage of Temporal Depthwise Separable Convolutions (TDSCs). Next, we reduce the number of fully connected layers using the Global Average Pooling (GAP) technique. The resulting proposed architecture is called Squeezed Very Deep Convolutional Neural Networks (SVDCNN).

\paragraph{Temporal Depthwise Separable Convolutions (TDSCs)}
The use of TDSCs over standard convolutions allowed reducing the number of parameters without relevant accuracy loss \cite{howard2017mobilenets}. TDSCs work decompounding the standard convolution into two parts: Depthwise and Pointwise. The first one is responsible for applying a convolutional filter to each channel of the input at a time. For an image input, one possibility of channels are the RGB components, whereas in a text input the dimensions of the embedding can be used instead. For both cases mentioned above, the result is one feature map by channel. The second convolution unifies the generated feature maps successively applying 1x1 convolutions so that the target amount of feature maps can be achieved.

\begin{figure}[htbp]
\begin{minipage}[t]{.5\linewidth}
\centering
\subfloat[]{\label{main:a}\includegraphics[scale=.105]{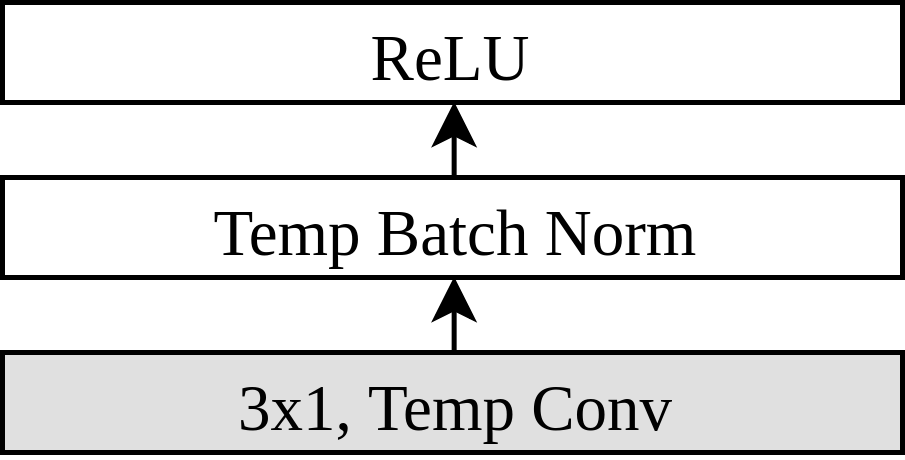}}
\end{minipage}%
\begin{minipage}[t]{.5\linewidth}
\centering
\subfloat[]{\label{main:b}\includegraphics[scale=.105]{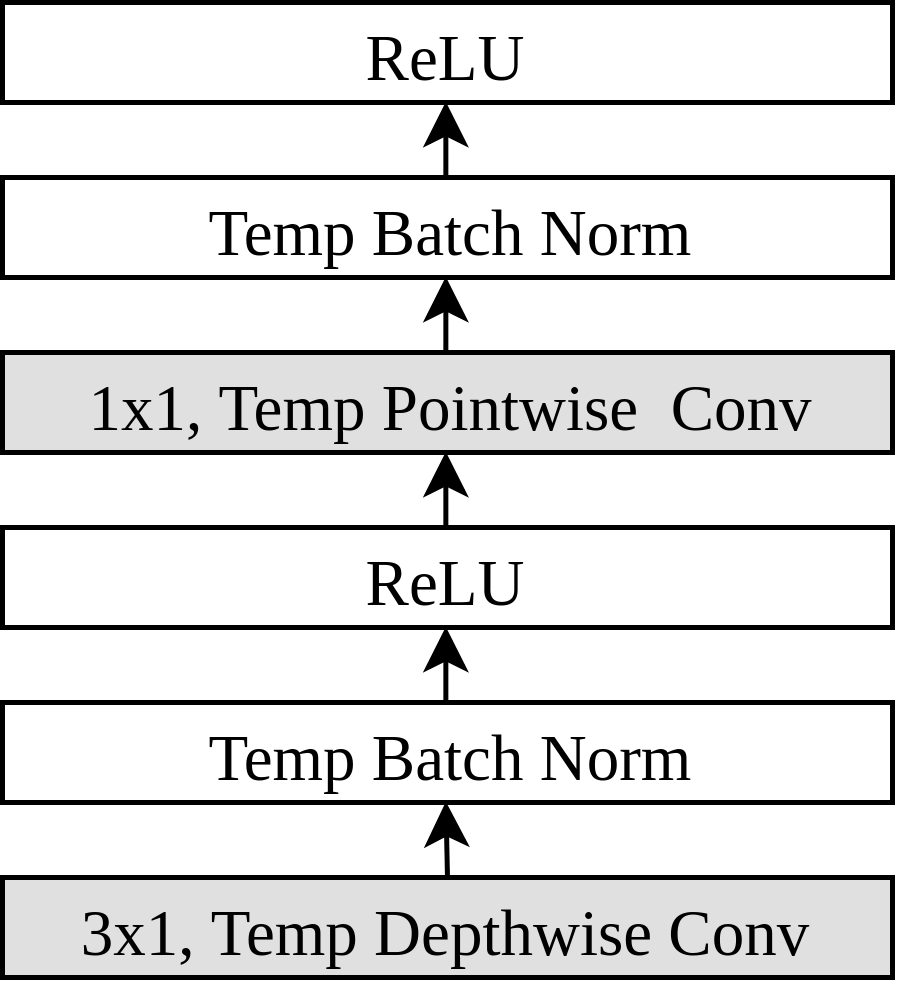}}
\end{minipage}
\caption{a) Temporal Standard Convolution;\\b) Temporal Depthwise Separable Convolution.}
\label{fig:main}
\end{figure}

\newpage

TDSCs are DSCs which work with one-dimensional convolutions. Although DSCs hold verified results in image classification networks, the usage of its temporal version for text related tasks is less explored. Fig. 2a presents the architecture of a temporal standard convolution while Fig. 2b presents the TDSC.

For a more formal definition, let  $P_{tsc}$ be the number of parameters of a temporal standard convolution, where  \textit{In} and \textit{Out} are the numbers of Input and Output channels respectively, and $D_k$  is the kernel size:

\begin{equation}
    P_{tsc} =  In \ast Out \ast D_k
\end{equation}

Alternatively, a TDSC achieves fewer parameters ($P_{tdsc}$): 

\begin{equation}
    P_{tdsc} = In \ast Dk  + In \ast Out 
\end{equation}

In the VDCNN model, one convolutional block is composed of two temporal standard convolutional layers. The first one doubles the number of feature maps while the second keeps the same value received as input. Besides, each convolutional layer is followed by a Batch Normalization and a ReLU layers. In our model, we proposed changing the temporal standard convolutions by TDSCs.

\begin{figure}[htbp]
\begin{minipage}[t]{.5\linewidth}
\centering
\subfloat[]{\label{main:a}\includegraphics[scale=.10]{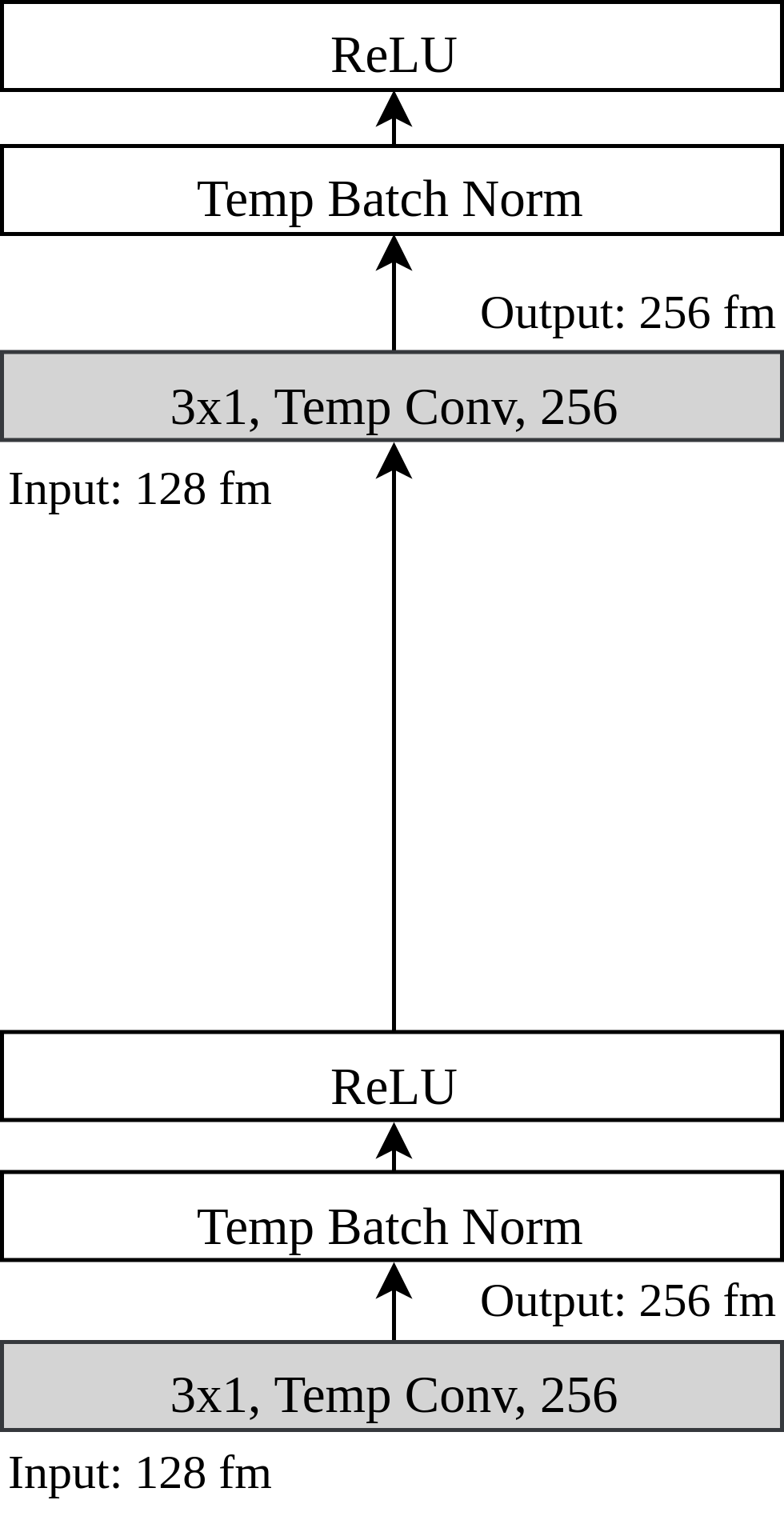}}
\end{minipage}%
\begin{minipage}[t]{.5\linewidth}
\centering
\subfloat[]{\label{main:b}\includegraphics[scale=.10]{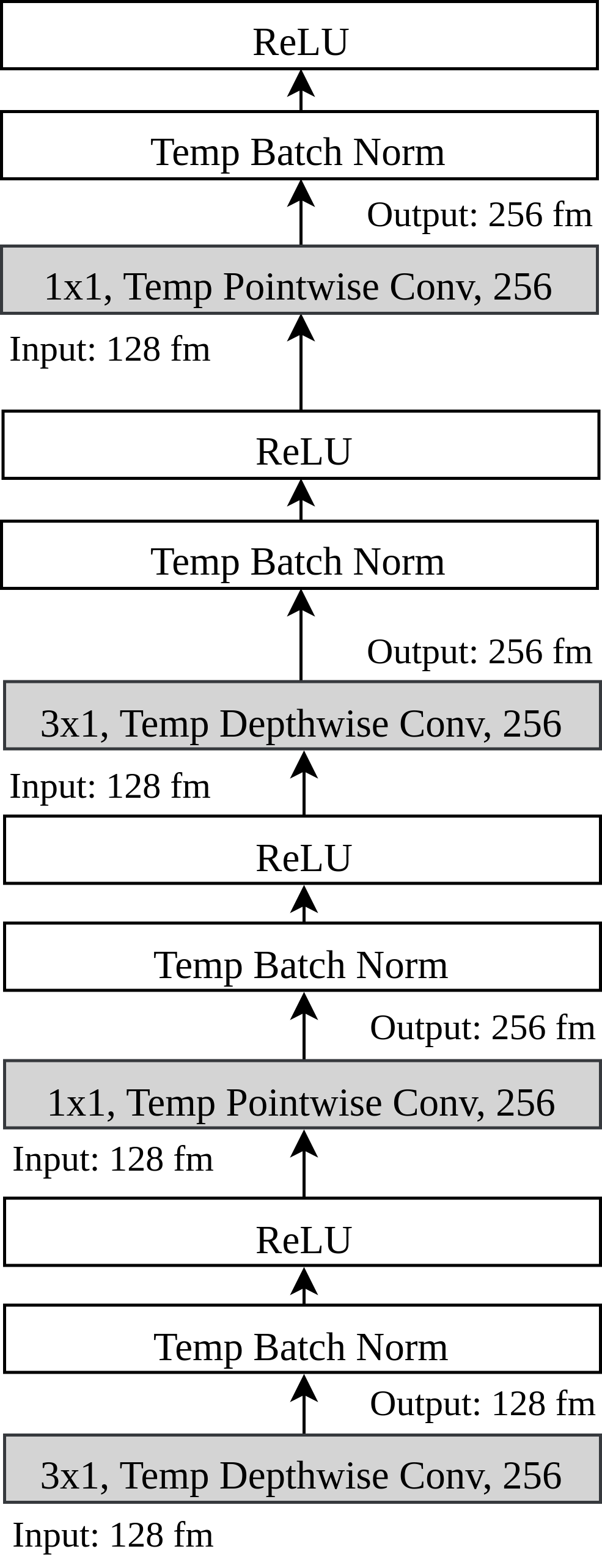}}
\end{minipage}
\caption{a) Standard convolutional block of the VDCNN;\\b) Modified convolutional block of the SVDCNN.}
\label{fig:main}
\end{figure}

Fig. 3 presents the standard convolutional block on the left and the proposed convolutional block using TDSC on the right. The pattern used in the figure for the convolutional layers is the following: "Kernel Size, Conv type, Output  Feature Maps"; as a brief example consider "3x1, Temporal Conv, 256", which means a Temporal Convolution with kernel size 3 and 256 feature maps as output. From Equation 1, we have the number of parameters of the original convolutional block  ($P_{convblock}$) as follows:


\begin{equation}
    P_{convblock} =  In \ast Out \ast 3 + Out \ast Out \ast 3 
\end{equation}

Moreover, from equation 2, the number of parameters of the proposed convolutional block ($P_{convblock-tdsc}$) that uses TDSC being:

\begin{equation}
    P_{convblock-tdsc} = In \ast 3 + In \ast Out + Out \ast 3  + Out \ast Out
\end{equation}

For illustration, following the same characteristics of Fig. 3, consider that the number of input channels In is equal to 128 and the number of output channels Out is equal to 256. Our proposed approach accumulates a total of 99,456 parameters. In contrast, there are 294,912 parameters in the original convolutional block. The use of TDSC yields a reduction of 66.28\% in the network size.

Lastly, since each standard temporal convolution turns into two  (Depthwise and Pointwise), the number of convolutions per convolutional block has doubled. Nevertheless, these two convolutions work as one because it is not possible to use them separately keeping the same propose. In this way, we count them as one layer in the network depth. This decision holds the provided depth architectures the same as the VDCNN model summarized in Table \ref{depth_vdcnn}, contributing to a proper comparison between the models. 

\paragraph{Global Average Pooling (GAP)}
The VDCNN model uses a $k$-max pooling layer $(k=8)$ followed by three fully connected (FC) layers to perform the classification task (Fig. 4a). Although this approach is the traditional architecture choice for text classification CNNs, it introduces a significant number of parameter in the network. The resulting number of the FC layers parameters ($P_{fc}$) aforementioned is presented below, for a problem with four target classes:

\begin{equation}
\begin{aligned}
     &  P_{fc} = 512 \ast k \ast 2,048 + 2,048 \ast 2,048 + 2,048 \ast 4 \\
    &  P_{fc} = 12,591,104
\end{aligned}
\end{equation}

\begin{figure}[htbp]
\begin{minipage}{.5\linewidth}
\centering
\subfloat[]{\label{main:a}\includegraphics[scale=.105]{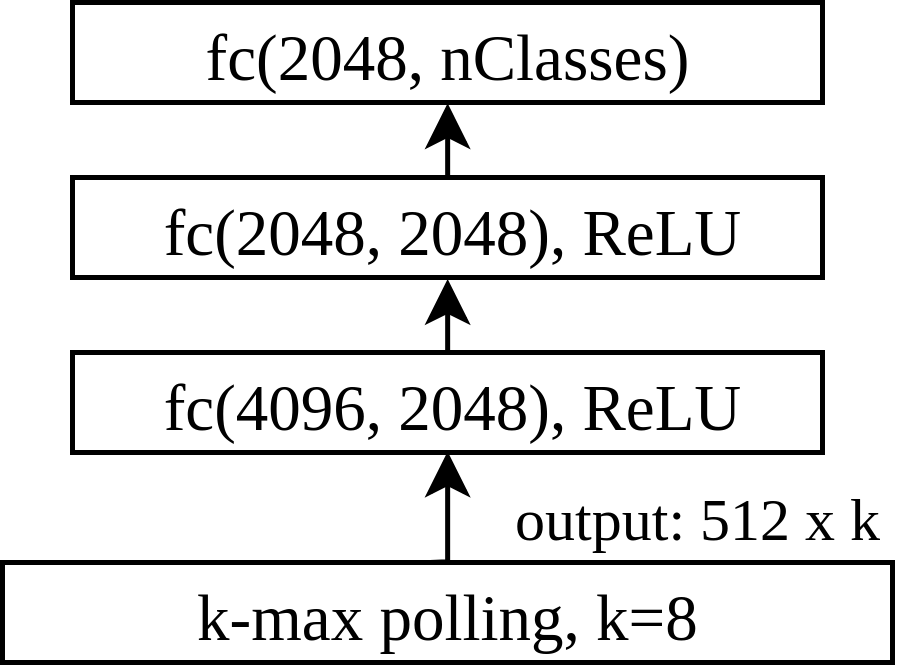}}
\end{minipage}%
\begin{minipage}{.5\linewidth}
\subfloat[]{\label{main:b}\includegraphics[scale=.105]{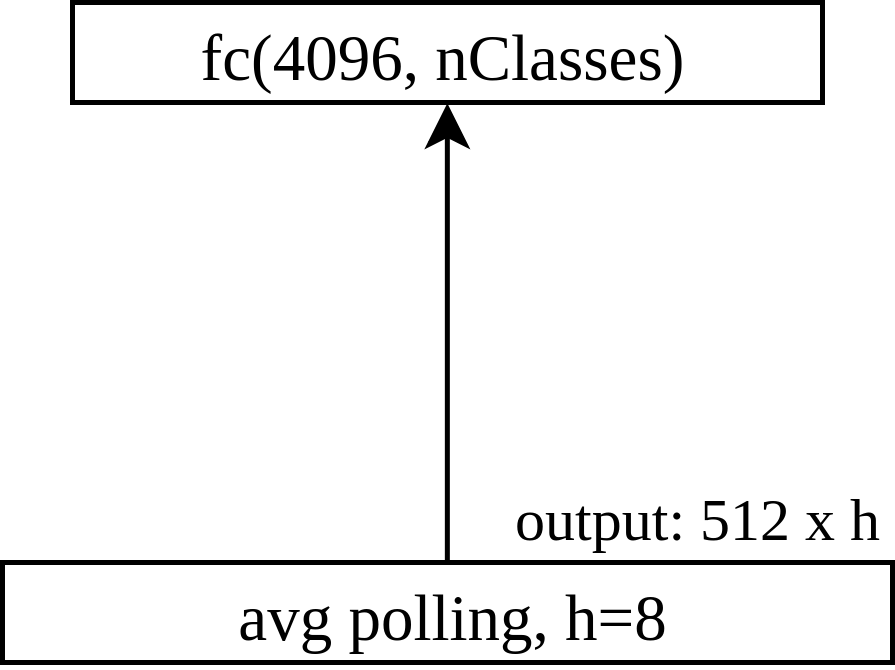}}
\end{minipage}
\caption{a) VDCNN classification layers;\\b) SVDCNN classification layers.}
\label{fig:main}
\end{figure}

\newpage

Instead of maintaining these fully connected layers, we directly aggregate the output of the last convolutional block through the usage of an average pooling layer. This method, known as Global Average Pooling, contributes substantially to the parameters reduction without degrading the network accuracy significantly \cite{lin2013network}. The number of resulting feature maps given by the average pooling layer was the same as the original $k$-max pooling layer $(k = h = 8)$.  Fig. 4b presents this proposed modification. The number of parameters obtained by the usage of GAP ($P_{gap}$) is revealed as follows:

\begin{equation}
\begin{aligned}
     & P_{gap} = 4,096 \ast 4 \\
    & P_{gap} = 16,384
\end{aligned}
\end{equation}

Our proposed approach accumulates a total of 16,384 parameters. In contrast, there are 12,591,104 parameters in the original classification method. The use of GAP yields a reduction of 99.86\%.

\section{Experiments}

The experiment goal is to investigate the impact of modifying the convolutional block of VDCNN to TDSCs and using GAP instead of the original fully connected layers. We evaluate Char-CNN, VDCNN, and SVDCNN according to the number of parameters, storage size, inference time and accuracy. The source code of the proposed model is available in the GitHub repository SVDCNN\footnote[1]{Link: \url{https://github.com/lazarotm/SVDCNN}}

The original VDCNN paper reported the number of parameters of the convolutional layers, in which we reproduce in this article. For SVDCNN and Char-CNN, we calculated the abovementioned number from the network architecture implemented in PyTorch. As for the FC layer's parameters, the number is obtained as the summation of the product of the input and output size of each FC layer for each CNN.




Considering the network parameters $P$ and assuming that one float number on Cuda environment takes 4 bytes, we can calculate the network storage in megabytes, for all the models, as follows:

\begin{equation}
     S = P \ast 4 \div 1,024^{2}
\end{equation}

\begin{table}[htpb]
\caption{Datasets used in experiments}
\ra{1.3}
\begin{center}
\begin{tabular}{@{}lllll@{}}\toprule
Dataset          & \#Train & \#Test & \#Classes & Classification Task  \\  \midrule
AG's News        & 120k  & 7.6k   & 4   & News categorization \\
Yelp Polarity & 560k & 38k  & 2  & Sentiment analysis  \\
Yelp Full & 650k & 50k  & 5  & Sentiment analysis  \\
\bottomrule
\end{tabular}
\end{center}
\label{datasets}
\end{table}

Regarding the inference time, its average and standard deviation were calculated as the time to predict one instance of the AG's News dataset throughout 1,000 repetitions. 

The SVDCNN experimental settings are similar to the original VDCNN paper, using the same dictionary and the same embedding size of 16 \cite{conneau2016very}.  The training is also performed with SGD, utilizing size batch of 64, with a maximum of 100 epochs. We use an initial learning rate of 0.01, a momentum of 0.9 and a weight decay of 0.001.  All the experiments were performed on an NVIDIA GTX 1060 GPU + Intel Core i7 4770s CPU.

The model's performance is evaluated on three large-scale public datasets also used by Zhang et al. \cite{zhang2015character} in the introduction of Char-CNN and VDCNN models. Table~\ref{datasets} presents the details of the utilized datasets: AG's News, Yelp Polarity and Yelp Full.

\section{Results}

Table~\ref{comparison_results} presents the number of parameters, storage size, and accuracy for the SVDCNN, VDCNN, and Char-CNN in all datasets. The use of TDSCs promoted a significant reduction in convolutional parameters compared to VDCNN. For the most in-depth network evaluated, which contains 29 convolutional layers (depth 29), the number of parameters of these convolutional layers had a reduction of 66.08\%, from 4.6 to 1.56 million parameters. This quantity is slightly larger than the one obtained from the Char-CNN, 1.40 million parameters, but this network has only six convolutional layers (depth 6).

The network reduction obtained by the GAP is even more representative since both compared models use three FC layers for their classification tasks. Considering a dataset with four target classes, and comparing SVDCNN with VDCNN, the number of parameters of the FC layers has passed from 12.59 to 0.02 million parameters, representing a reduction of 99.84\%.  Following with the same comparison, but to Char-CNN, the proposed model is 99.82\% smaller, 0.02 against 11.36 million of FC parameters.

The reduction of the total parameters impacts directly on the storage size of the networks. While our most in-depth model (29) occupies only 6MB, VDCNN with the same depth occupies 64.16MB of storage. Likewise, Char-CNN (which has depth 6) occupies 43.25MB. This reduction is a significant result because many embedded platforms have several memory constraints. For example, FPGAs often have less than 10MB of on-chip memory and no off-chip memory or storage \cite{howard2017mobilenets}.

Regarding accuracy results, usually, a model with such parameter reduction should present some loss of accuracy in comparison to the original model. Nevertheless, the performance difference between VDCNN and SVDCNN models varies between 0.4 and 1.3\%, which is pretty modest considering the parameters and storage size reduction aforementioned. In Table \ref{comparison_results}, it is possible to see the accuracy scores obtained by the compared models. Another two fundamental results obtained are a) The base property of VDCNN model is preserved on its squeezed model: the performance still increasing up with the depth and b) The performance evaluated for the most extensive dataset, i.e., Yelp Review (62.30\%), still overcomes the accuracy of the Char-CNN model (62.05\%).

Deep learning processing architecture has the property of being high parallelizable; it is expected smaller latencies when performing inferences in hardware with high parallelization power. Despite this property, the model ability to use all hardware parallel potential available also depends on the network architecture. The more parameters per layers, the more parallelizable a model tends to be, while the increase of the depth gets the opposite result. Another natural comprehension fact is if a model has few parameters, there exists less content to be processed, and then we have a faster inference time. 

Concerning mobile devices, the presence of dedicated hardware for deep learning is not entirely feasible. This hardware usually requires more energy and dissipates more heat, two undesirable features for a mobile platform. Therefore, obtaining fewer inference times, even out of environments with high parallelization capabilities, is a pretty desirable characteristic for a model designed to work on mobile platforms. The latency ratio between CPU and GPU inference times indicates how undependable of dedicated hardware a model is, with higher values meaning more independence. 

The inference times obtained for the three models compared are available in Table \ref{time_results}. As explained in Section IV a), each convolutional layer of the convolutional blocks was substituted by two convolutions. This change could impact the inference time negatively, but the significant parameter reduction allows the SVDCNN to obtain better results than the VDCNN model.

\begin{table}[!b]
\caption{Time results for AG's News dataset}
\label{time_results}
\ra{1.3}
\begin{center}
\begin{tabular}{@{}rrrr@{}}\toprule
& \multicolumn{3}{c}{Inference Time} \\
\cmidrule{2-4}
& GPU & CPU & Ratio \\ \midrule
\textbf{SVDCNN}\\
9 & $5.53\text{ms}\pm0.16$ & $25.88\text{ms}\pm0.52$ & 0.21 \\
17 & $9.84\text{ms}\pm0.28$ & $47.80\text{ms}\pm1.01$ & 0.21 \\
29 & $15.14\text{ms}\pm0.44$ & $74.03\text{ms}\pm1.15$ & 0.20 \\
\textbf{VDCNN}\\
9 & $4.48\text{ms}\pm0.19$ & $29.13\text{ms}\pm0.87$ & 0.15 \\
17 & $7.08\text{ms}\pm0.20$ & $48.05\text{ms}\pm1.26$ & 0.15 \\
29 & $10.26\text{ms}\pm0.26$ & $65.80\text{ms}\pm1.51$ & 0.16 \\
\textbf{Char-CNN}\\
6 & $10.32\text{ms}\pm0.43$ & $313.53\text{ms}\pm4.97$ & 0.03 \\
\bottomrule
\end{tabular}
\end{center}
\end{table}

\begin{table*}[htpb]\centering
\caption{Number of parameters, storage and accuracy results for all evaluated CNNs}
\label{comparison_results}
\ra{1.3}
\begin{tabular}{@{}lrrrcrrrcr@{}}\toprule
& \multicolumn{3}{c}{SVDCNN} & \phantom{abc}& \multicolumn{3}{c}{VDCNN} &
\phantom{abc} & Char-CNN\\
\cmidrule{2-4} \cmidrule{6-8}  \cmidrule{10-10}
& 9 & 17 & 29 && 9 & 17 & 29 && 6\\ \midrule
\textbf{Parameters}\\
\#Conv Params [M] & 0.71 & 1.43 & 1.56 && 2.20 & 4.40 & 4.60 && 1.37\\
\#FC Params [M] & 0.02 & 0.02 & 0.02 && 12.59 & 12.59 & 12.59 && 11.34\\
\#Total Params [M] & 0.73 & 1.45 & 1.58 && 14.79 & 16.99 & 17.19 && 12.71\vspace{5px}
\\
\textbf{Storage}\\
Storage Size [MB] & 2.80 & 5.52 & 6.03 && 54.75 & 62.74 & 64.16 && 43.25\vspace{5px}
\\
\textbf{Accuracy}\\
Ag News & 90.13& 90.43&90.55& & 90.83& 91.12& 91.27&& 92.36\\
Yelp Polarity &94.99& 95.04& 95.26&& 95.12& 95.50& 95.72&& 95.64\\ 
Yelp Full &61.97& 63.00& 63.20&& 63.27& 63.93& 64.26&& 62.05\vspace{5px}
\\
\bottomrule
\end{tabular}
\end{table*}

\noindent The CPU inference time obtained by the proposed model was smaller than the base model for the depth 9 (25.88ms against 29,13ms) and depth 17 (47.80ms against 48.05ms), while the Ratio was higher for all depths (0.20 against 0.15 in average). These results, as explained above, are pretty significant for mobile platforms. Looking to Char-CNN, this model got notably inferior results compared to the proposed method, with 313.53ms of CPU inference time and Ratio of 0.03.

\section{Conclusion}
In this paper, we presented a squeezed version of the VDCNN model considering the number of parameters and size. The new model proprieties became it feasible for mobile platforms. To achieve this goal, we analyzed the impact of including Temporal Depthwise Separable Convolutions and a Global Average Pooling layer in a very deep convolutional neural network for text classification. The SVDCNN model reduces about 92.45\% the number of parameters and storage size while presents an inference time ratio (CPU/GPU), 31.94\% higher.  

For future works, we plan to evaluate other techniques able to reduce storage size, such as model compression. Moreover, the model accuracy over even more massive datasets will be evaluated as well as the efficiency of its depth 49 configuration.

\section*{Acknowledgment}
We would like to thank FACEPE and CNPq (Brazilian research agencies) for financial support.


\bibliographystyle{IEEEtran}  
\bibliography{IEEEtran/sample}  

\end{document}